\documentclass[sigconf]{acmart}

\setcopyright{iw3c2w3}
\usepackage{booktabs}

\newcommand{\boilernet}{\textsc{BoilerNet}}
\newcommand{\wtt}{\textsc{Web2Text}}
\newcommand{\bp}{\textsc{BoilerPipe}}
\newcommand{\moz}{\textsc{Readability.js}}
\newcommand{\bte}{\textsc{BTE}}
\newcommand{\cetd}{\textsc{CETD}}

\begin{document}

\copyrightyear{2020}
\acmYear{2020} 
\acmConference[WWW '20 Companion]{Companion Proceedings of the Web Conference 2020}{April 20--24, 2020}{Taipei, Taiwan}
\acmBooktitle{Companion Proceedings of the Web Conference 2020 (WWW '20 Companion), April 20--24, 2020, Taipei, Taiwan}
\acmPrice{}
\acmDOI{10.1145/3366424.3383547}
\acmISBN{978-1-4503-7024-0/20/04}

\title{Boilerplate Removal using a Neural Sequence Labeling Model}

\author{Jurek Leonhardt}
\affiliation{
  \institution{L3S Research Center}
  \city{Hannover}
  \country{Germany}
}
\email{leonhardt@L3S.de}

\author{Avishek Anand}
\affiliation{
  \institution{L3S Research Center}
  \city{Hannover}
  \country{Germany}
}
\email{anand@L3S.de}

\author{Megha Khosla}
\affiliation{
  \institution{L3S Research Center}
  \city{Hannover}
  \country{Germany}
}
\email{khosla@L3S.de}

\begin{abstract}
The extraction of main content from web pages is an important task for numerous applications, ranging from usability aspects, like \textit{reader views} for news articles in web browsers, to information retrieval or natural language processing. Existing approaches are lacking as they rely on large amounts of hand-crafted features for classification. This results in models that are tailored to a specific distribution of web pages, e.g.\ from a certain time frame, but lack in generalization power. We propose a neural sequence labeling model that does not rely on any hand-crafted features but takes only the HTML tags and words that appear in a web page as input. This allows us to present a browser extension which highlights the content of arbitrary web pages directly within the browser using our model. In addition, we create a new, more current dataset to show that our model is able to adapt to changes in the structure of web pages and outperform the state-of-the-art model.
\end{abstract}

\maketitle
\section{Introduction}
Web pages are rich sources of information but are also intertwined inextricably with ads, banners and other CMS boilerplate. Extracting the primary content from a Web page is an important low level task with strong implications to retrieval models~\cite{vogels2018web2text} and other content extraction tasks. 

In this paper we focus on the task of boilerplate removal or isolating the primary informational content of a web page. The problem is a well studied one with approaches ranging from classical rule-based ones to modern approaches that model it as a supervised machine learning problem. Most of the recent approaches rely on building large set of features based on commonly observed domain-knowledge of rules. \citet{Gupta2003} and \citet{wu2015automatic} rely on DOM-based features or structure of web pages, \bp~\cite{kohlschutter2010boilerplate} and \wtt~\cite{vogels2018web2text} rely on text-based features, and \citet{cai2004block} use vision-based page segmentation approaches. Most of these approaches exploit a crucial aspect that is common to most web pages, i.e.\ there is an inherent locality of relevant content~\cite{finn2001fact}. In other words, relevant content tends to be rendered closely together in a web page.

There are two major drawbacks of the past approaches. First, the locality of rendering effect is inherently hard to model and requires a large number of features and heuristic post-processing procedures as exemplified in the previous approaches that use DOM trees, text or a combination of both. For example the state-of-art approach~\wtt{} uses $128$ features. Second, the availability of training data for the boilerplate removal task is limited and for maintaining the effectiveness of these models on evolving web pages they have to be re-trained. With human training labels being at a premium, this limited training data is a common issue that plagues all existing models.

\begin{figure}
	\centering
	\includegraphics[width=\linewidth]{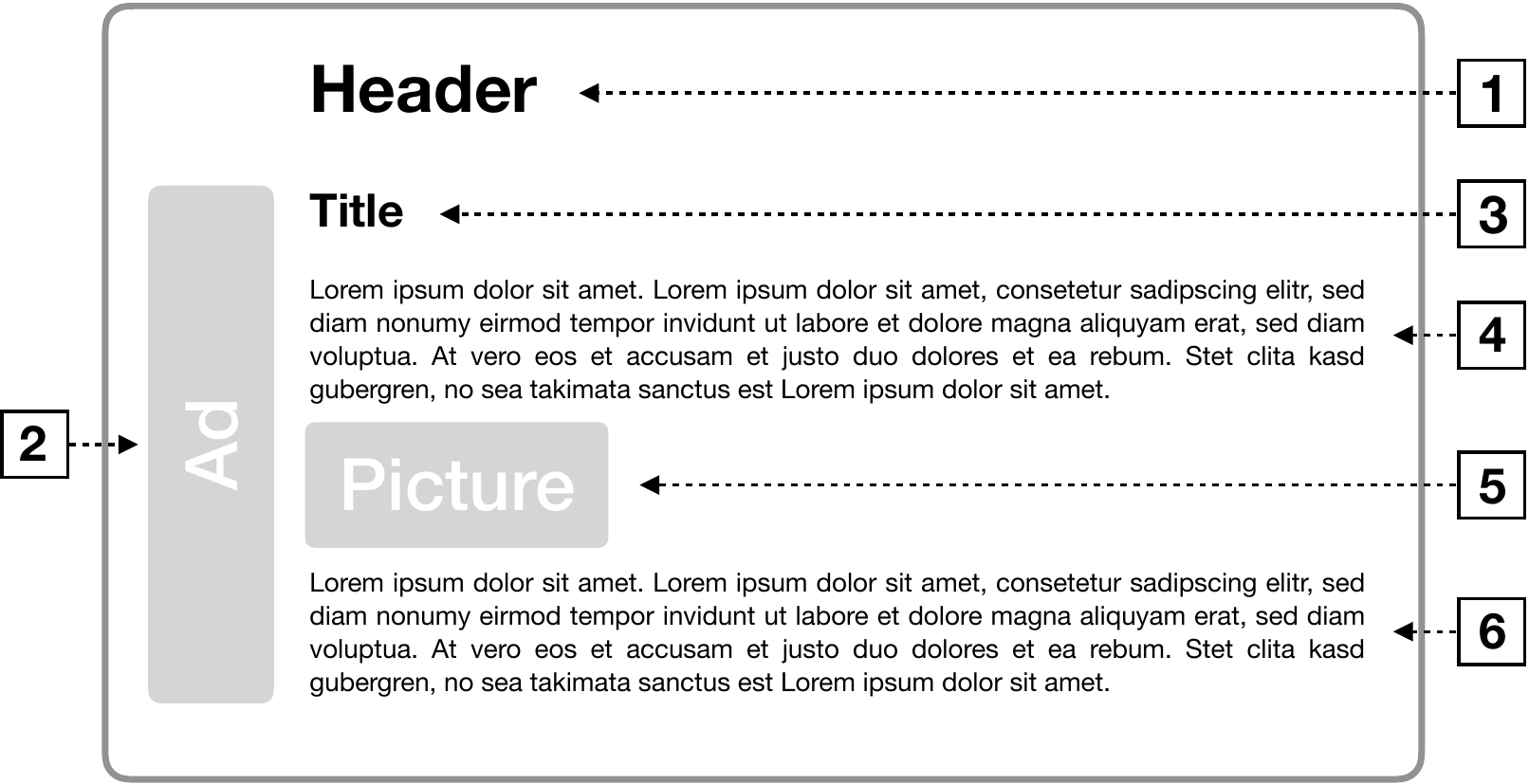}
	\caption{A common example layout that is found e.g.\ in blogs. This web page can be represented as a sequence, the elements of which are illustrated by the numbers corresponding to the page elements. The order of the sequence is determined by the order of the elements within the DOM tree.}
	\label{fig:layout}
\end{figure}

In this paper we propose \boilernet{} as an automatic feature learning approach that does not rely on expensive feature engineering on web pages. Unlike the locality of rendering, there is also a \emph{locality of authoring} in web pages. Specifically, relevant content also tends to be authored in contiguous segments, as illustrated in Figure \ref{fig:layout}. This observation greatly simplifies the modelling task where we are not bound to inferring the locality of content from the visual features or raw text. Rather, we work on the raw HTML input and model the input web page as a sequence of candidate segments (text blocks, i.e.\ leaf nodes in the DOM tree). Assuming that the content is authored in a sequential fashion, we now pose the problem of detecting boilerplate as a sequence learning task. This eradicates the utility of any computationally expensive feature extraction and post-processing procedures. We further show that our model is able to perform and generalize well even with a low number of training instances. We perform extensive experimental evaluation with standard benchmarks as well as a new dataset created by us. We observe that we obtain similar performance and sometimes outperform the existing feature-based approaches. In sum, we make the following contributions:
\begin{itemize}
	\item We propose \boilernet{} as an automatic feature learning approach that is based only on raw HTML pages. It is robust to heterogeneous genres, authoring practices and learns effectively in low training data regimes.
	\item We conduct empirical evaluation to show that \boilernet{} performs on par with or better than existing state-of-the-art approaches.
	\item We provide an interactive demonstration in the form of a browser extension.
\end{itemize}

\section{Related Work}
Existing boilerplate removal approaches are either $(1)$ handcrafted rules or tools targeted to extract content from web pages which were observed to possess certain structural and textual properties, or $(2)$ machine learning approaches using hand-crafted (textual, structural, visual etc.) features to separate content from boilerplate.

For instance, Body Text Extraction (\bte)~\cite{finn2001fact} uses the observation that the main content contains longer paragraphs of uninterrupted text and marks the largest contiguous text area with the least amount of HTML tags as content. \cite{Gupta2003} applies various heuristic-based filters to remove images, advertisements etc.\ from the DOM tree representation of the web page. \cetd~\cite{sun2011dom} exploits the design principles behind text and noise. Another line of work consists of template detection algorithms \cite{Bar-Yossef2002,Lin2002,Yi2003,chakrabarti2007page} which utilize collections of web pages, usually from the same site, to learn the common template structure.

Among the other machine learning based approaches, \cite{Pasternack} proposes a method utilizing \textit{maximum subsequence segmentation} to extract the text of articles from HTML documents using tags and trigram features. \bp~\cite{kohlschutter2010boilerplate} analyzes hand-crafted text features, namely the number of words and link density, to distinguish main content from other parts of information from news article web pages. \cite{Wang2009} learns a template-independent wrapper using a small number of labeled news pages from a single site and features dedicated to news titles and bodies. \cite{gibson2007adaptive} and \cite{spousta2008victor} employ sequence labeling approaches which consider a web document as a sequence of some appropriately sized units or blocks and the task is to categorize each block as \emph{content} or \emph{not content}. Both of these works use Conditional Random Fields (CRFs) models using various hand-crafted textual and structural features. \wtt~\cite{vogels2018web2text} employs two separate convolutional neural networks which operate on a very large number of hand-crafted features for each text block, yielding unary and pairwise potentials (probabilities), for classifying each block and pair of adjacent blocks respectively followed by maximization of joint probabilities during inference. \cite{wu2015automatic} formulates the actual content identifying problem as a DOM tree node selection problem and trains a machine learning model using features like fonts, links, position and others.

\section{BoilerNet}
We model the problem of boilerplate removal or content extraction as a \emph{sequence labeling} problem where the web page is divided into text blocks, all of which are then individually classified as either content or boilerplate. Each text block is represented as a vector which encodes both the text as well as all of its ancestral HTML tags. The order of the input sequence is determined by the order of the corresponding text blocks within the original HTML file. Our hypothesis is that the order of text blocks in a web page encodes important information about their type, i.e.\ content or boilerplate, as the placement is determined by the authoring style. For example, in blogs or news sites, we expect the content of an article page to be bunched near the center and surrounded by ads and navigational elements.

\subsection{Input Representation} In order to obtain the sequences to be used as inputs for our model, we divide every web page into a sequence of text blocks. A text block can only appear as a leaf node in the DOM tree and has no associated HTML tag. However, a piece of text that appears connected to the human reader may be separated by HTML tags, e.g.\ \texttt{<p>[A]<strong>[B]</strong>[C]</p>} contains the text blocks \texttt{[A]}, \texttt{[B]} and \texttt{[C]}, divided by a \texttt{<strong>} tag.

Each block is represented by a $d$-dimensional vector. The first $k$ elements of the vector are used to encode all parent nodes of the leaf up to the root. Each index is associated with a specific HTML tag, and the numbers encode how many of the corresponding HTML tags appear in the path from the root node to the leaf. In the same fashion, the remaining $l$ items encode the words that appear in the text block. In practice, we add another two dimensions for out-of-vocabulary placeholders to the vector in order to handle unknown HTML tags and unknown words, and only consider the $k$ most common HTML tags and the $l$ most common words.

\begin{figure}
	\centering
	\includegraphics[width=\linewidth]{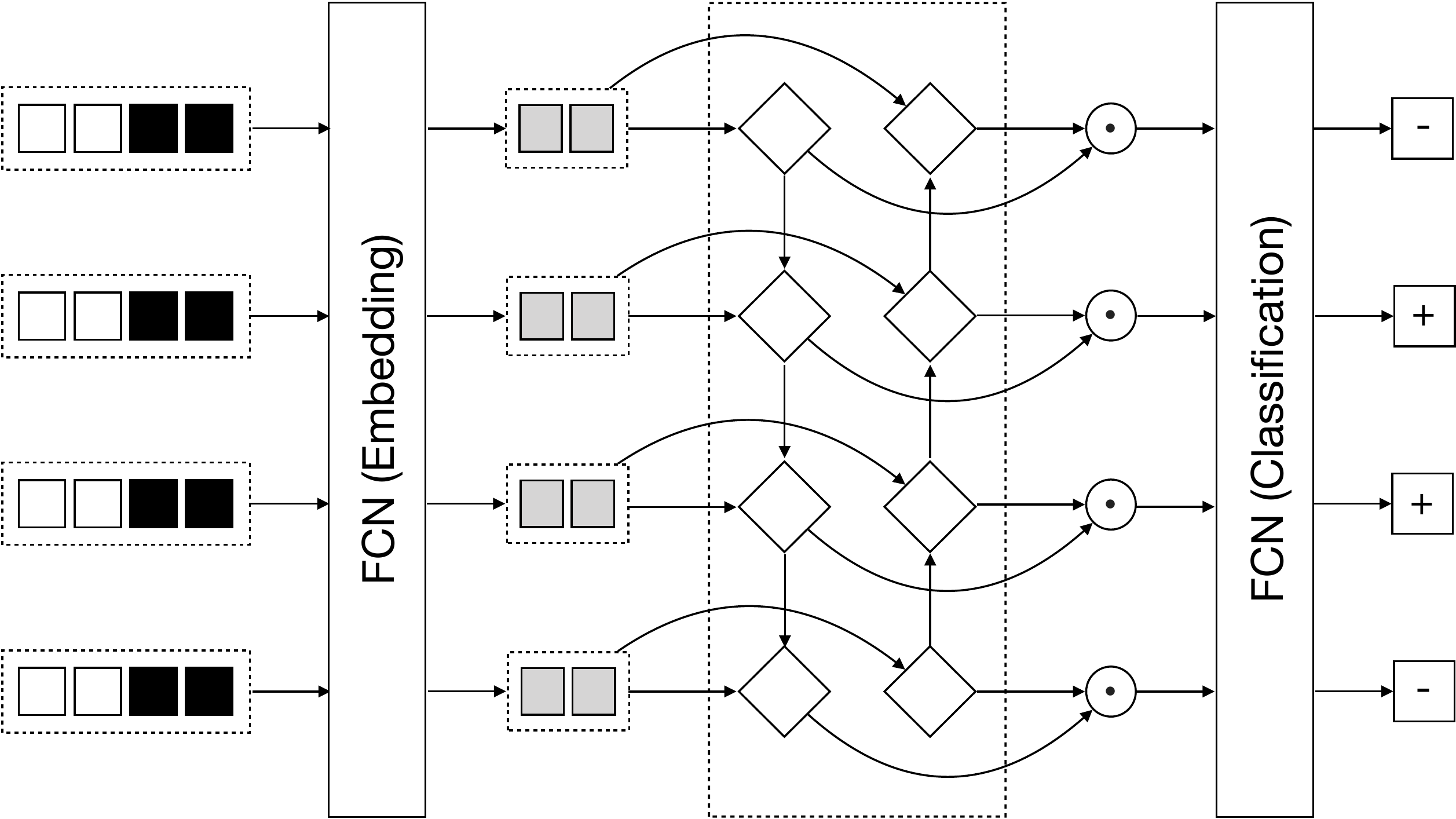}
	\caption{The \boilernet{} architecture with one LSTM layer. The input is a web page represented as a sequence of text blocks. We represent each text block using a sparse vector which encodes the HTML tags (white) and words (black). These input vectors are then transformed into lower-dimensional dense vectors using an embedding layer. We classify each element of the sequence.}
	\label{fig:model}
\end{figure}

\subsection{Sequence Labeling}
The \boilernet{} architecture is based on bidirectional LSTMs which are able to learn complex non-local dependencies in sequence models. The elements in the input sequence are projected to $m$-dimensional dense vectors using a fully-connected layer $D \in \mathbb{R}^{d \times m}$. The dense representations are then fed into a number of consecutive many-to-many bidrectional LSTM layers. In each of these LSTM layers, given an input $(x_1, x_2, ..., x_n)$, the forward LSTM yields the output $(y_1^f, ..., y_n^f)$ while the backward LSTM receives the reversed input sequence and yields the output $(y_1^b, ..., y_n^b)$. Both LSTMs use individual sets of parameters $W^f$ and $W^b$. The final $\ell$-dimensional representation of a text block is obtained by concatenating the respective forward and backward outputs. Finally, the concatenated outputs of the last LSTM layer are squashed into $1$-dimensional vectors using a fully connected layer $V \in \mathbb{R}^{\ell \times 1}$ with a sigmoid activation function to infer the final classification probabilities. We use binary cross entropy to train the model. The architecture is shown in Figure \ref{fig:model}.

\subsection{Issues in Boilerplate Removal Models}
One of the main issues in employing machine learning models to classify content and boilerplate is the scarcity of the labeled data. Annotating content in web pages is a tedious task and requires a large investment of time, mainly because of the growing size of web pages and non-standard use of HTML. We initially experimented with a \emph{weak supervision}~\cite{dehghani2017fidelity} strategy to deal with this problem, which resulted in no significant improvements. We therefore conduct our experiments with a low number of training examples to show that our model is able to generalize well despite limited training data.

\section{Experiments}
We evaluate \boilernet{} against existing approaches using two datasets: CleanEval~\cite{baroni2008cleaneval} and GoogleTrends-2017. CleanEval (published in 2007) is a collection of arbitrary websites. In order to evaluate the generalizability of our model to newer web pages, we manually created GoogleTrends-2017 as described in Section \ref{sec:data_prep}.

We compare our approach with other machine learning-based as well as rule- and heuristic-based methods. \wtt~\cite{vogels2018web2text} relies on a large number of hand-crafted features and uses convolutional neural networks to classify page elements. The drawback of this approach is that these features might not generalize well or in some cases even be invalid, for example across languages or longer time periods. \bp{}~\cite{kohlschutter2010boilerplate} is a rule-based system using textual features which was trained solely on news articles.
Finally, \moz{}\footnote{\url{https://github.com/mozilla/readability}} is the open source implementation of Mozilla's \emph{reader view} feature in the Firefox browser. 

\subsection{Dataset Preparation}
\label{sec:data_prep}
Existing datasets for the evaluation of boilerplate removal and content extraction have weaknesses, rendering them suboptimal for evaluating modern approaches; firstly, the datasets are old (CleanEval was published in 2007, L3S-GN1 in 2010). Since then, the web has changed in many ways, for example in overall structure and technologies used, rendering existing datasets outdated. Secondly, some datasets lack diversity, as they contain only web pages of a single type. For example, L3S-GN1 is a news-only dataset.

For these reasons we created a new dataset based on the Google trends of 2017. We obtained the HTML files by retrieving the first $100$ results for each trending Google query from the year 2017.\footnote{\url{https://trends.google.com/trends/yis/2017/GLOBAL/}} From the resulting pool of websites we randomly sampled a set of $180$ documents and annotated them. To eliminate the problems mentioned above, we made sure to retrieve all our pages from current Google queries, as this ensures both variety and currentness of the data.


\subsection{Results and Discussion}
We conducted experiments on the CleanEval and GoogleTrends-2017 datasets to compare \boilernet{} with the baselines. Due to the scarcity of ground-truth data mentioned earlier, we intentionally use small numbers of training instances. This also has the advantage of giving us a larger testset. We also experimented with $5$-fold cross-validation, however this did not give us any significant improvements. Our final model after validation contains two bidirectional LSTM layers with $256$ hidden units each. The sparse input vectors are projected to $256$-dimensional embedding vectors. After the second LSTM layer, we apply dropout with a probability of $0.5$. We train each model for $50$ epochs with a batch size of $16$ and weighted binary cross entropy loss. We choose the best checkpoint based on the $F_1$ score on the validation set.

\begin{table}
\caption{The results on the CleanEval dataset with 55 training, 5 validation and 676 test instances.}
\setlength\tabcolsep{0pt} 
\renewcommand*{\arraystretch}{0.8}
\begin{tabular*}{\columnwidth}{l @{\extracolsep{\fill}} *{6}{c}}
    \toprule
                    & \multicolumn{3}{c}{Negative class}            & \multicolumn{3}{c}{Positive class}\\
                      \cmidrule{2-4}                                  \cmidrule{5-7}
                    & {$P$}         & {$R$}         & {$F_1$}       & {$P$}         & {$R$}         & {$F_1$}\\
    \midrule
    \wtt            & 0.82          & 0.76          & 0.79          & 0.84          & \textbf{0.89} & 0.86\\
    \wtt{} (unary)  & 0.80          & 0.78          & 0.79          & 0.85          & 0.87          & 0.86\\
    \bp             & 0.58          & \textbf{0.90} & 0.71          & \textbf{0.90} & 0.58          & 0.71\\
    \moz            & \textbf{0.85} & 0.76          & 0.81          & 0.82          & \textbf{0.89} & 0.85\\
    \midrule
    \boilernet      & 0.82          & 0.83          & \textbf{0.82} & 0.87          & 0.86          & \textbf{0.87}\\
    \bottomrule
\end{tabular*}
\label{tab:cleaneval}
\end{table}

\begin{table}
\caption{The results on the new GoogleTrends-2017 dataset with 50 training, 30 validation and 100 test instances.}
\setlength\tabcolsep{0pt} 
\renewcommand*{\arraystretch}{0.8}
\begin{tabular*}{\columnwidth}{l @{\extracolsep{\fill}} *{6}{c}}
    \toprule
                    & \multicolumn{3}{c}{Negative class}            & \multicolumn{3}{c}{Positive class}\\
                      \cmidrule{2-4}                                  \cmidrule{5-7}
                    & {$P$}         & {$R$}         & {$F_1$}       & {$P$}         & {$R$}         & {$F_1$}\\
    \midrule
    \wtt            & 0.88          & 0.89          & 0.88          & 0.69          & 0.67          & 0.68\\
    \wtt{} (unary)  & 0.91          & 0.85          & 0.88          & 0.67          & 0.77          & 0.71\\
    \bp             & 0.78          & \textbf{0.98} & 0.87          & \textbf{0.85} & 0.26          & 0.39\\
    \moz            & 0.84          & 0.85          & 0.84          & 0.53          & 0.52          & 0.53\\
    \midrule
    \boilernet      & \textbf{0.95} & 0.86          & \textbf{0.90} & 0.70          & \textbf{0.88} & \textbf{0.78}\\
    \bottomrule
\end{tabular*}
\label{tab:googletrends}
\end{table}

The results on the CleanEval dataset are shown in Table \ref{tab:cleaneval}. We used the original CleanEval split, i.e.\ $55$ training instances, $5$ validation instances and $676$ test instances. Unlike all other approaches, \boilernet{} is consistent in both classes, whereas the recall of \wtt{} drops in the negative class. The overall performance of \boilernet{} and \wtt{} is similar, which shows that our approach is able to learn the features of \wtt{} that were optimized for CleanEval. Table \ref{tab:googletrends} shows the results on the GoogleTrends-2017 dataset. Again, \boilernet{} is more consistent than the other approaches, outperforming them especially in the positive class. This supports our claim that hand-crafted features do not generalize well across longer periods of time. Moreover, while for other approaches it would be hard to detect boilerplate text that contains well formed sentences (e.g.\ copyright statements), in the case of raw input (like ours) we learn to predict such cases with higher accuracy.

\section{Demonstration}
\begin{figure}
	\centering
	\includegraphics[width=0.6\linewidth]{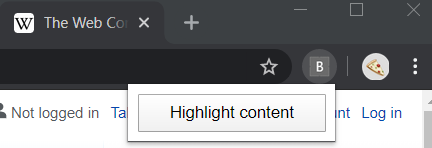}
	\caption{The user interface of the browser extension.}
	\label{fig:ui}
\end{figure}

\begin{figure}
	\centering
	\includegraphics[width=\linewidth]{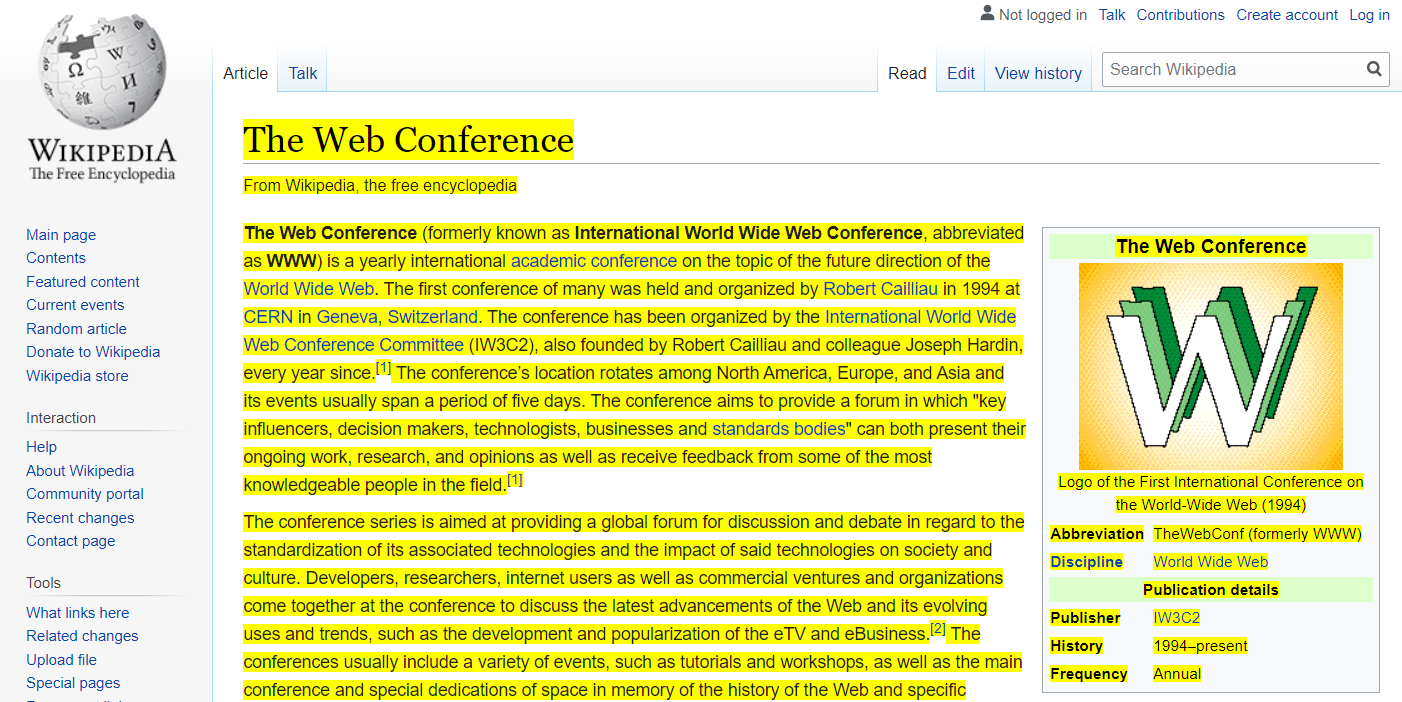}
	\caption{An example web page where the content has been highlighted by the \boilernet{} browser extension.}
	\label{fig:example}
\end{figure}

We demonstrate our system by providing an interactive browser extension that allows to user to highlight the content on an arbitrary web page with a single button click. The user interface is shown in Figure \ref{fig:ui}. After the user initiates the process, the extension pre-processes the current web page, loads a pre-trained \boilernet{} model and classifies each text element. Those elements which are classified as content are then highlighted directly within the active browser tab. Figure \ref{fig:example} shows an example web page where the content has been highlighted.

\subsection{Implementation Details}
Our implementation of the \boilernet{} model uses TensorFlow 2.0. The training (and evaluation) happens on a GPU. A trained model can then be loaded by our browser extension which uses TensorFlow.js, a JavaScript library compatible to TensorFlow.\footnote{\url{https://www.tensorflow.org/js}} The extension is self-contained, i.e.\ it handles all necessary pre-processing steps like the tokenization of text paragraphs. The browser extension, including a pre-trained model, is publicly available for download.\footnote{\url{https://github.com/mrjleo/boilernet/releases}}

\section{Conclusion}
We presented \boilernet{}, a novel, featureless approach for boilerplate removal from web pages using sequence labeling. We have shown that \boilernet{} can match the performance of state-of-the-art systems and outperform them on more current datasets, achieving an increase of $11\%$ in recall and $7\%$ in $F_1$ (positive class) over its competitors. We have also shown the benefits of modeling web pages as sequences of text blocks while preserving the order from the DOM tree. Additionally, we have shown that our approach requires only little training data to achieve good results. Our \boilernet{} implementation is publicly available, including our browser extension for demonstration purposes.\footnote{\url{https://github.com/mrjleo/boilernet}}

\section{Acknowledgements}
This work is partially funded by SoBigData++ (EU Horizon 2020 grant agreement no.\ 871042).

%
%



\bibliographystyle{ACM-Reference-Format}
\bibliography{bibliography}

\end{document}